\documentclass[dvipsnames]{article}

\usepackage{microtype}
\usepackage{graphicx}
\usepackage{booktabs} 

\usepackage{hyperref}

\usepackage[numbers, sort, compress]{natbib}

\usepackage[table]{xcolor}
\usepackage[utf8]{inputenc} 
\usepackage[T1]{fontenc}    
\usepackage{hyperref}       
\usepackage{url}            
\usepackage{booktabs}       
\usepackage{amsfonts}       
\usepackage{nicefrac}       
\usepackage{microtype}      
\usepackage{caption}
\usepackage{multicol}
\usepackage{multirow}
\DeclareCaptionType{equ}[][]

\usepackage{amsfonts}
\usepackage{amsmath, mathtools}
\usepackage{amsthm}
\usepackage{bm}

\usepackage{cleveref}
\usepackage{subcaption}
\usepackage{wrapfig}
\allowdisplaybreaks
\usepackage{booktabs} 
\usepackage{tabularx}
\usepackage[most]{tcolorbox}

\usepackage{comment}

\usepackage{tikz}
\usepackage{pgfplots}
\usetikzlibrary{shapes.geometric,arrows,arrows.meta, fit}

\newcommand{\bx}[0]{\mathbf{x}}

\newcommand{\liraforget}[0]{\texttt{LiRA-Forget}}
\newcommand{\ftheta}{f_{\theta(S)}}
\newcommand{\probdist}[0]{p_{S}}
\newcommand{\probretrain}[0]{p_{S \backslash S_f}}
\newcommand{\probunlearn}[0]{p_{\mathcal{U}}}

\makeatletter
\newcommand*{\addFileDependency}[1]{
  \typeout{(#1)}
  \@addtofilelist{#1}
  \IfFileExists{#1}{}{\typeout{No file #1.}}
}
\makeatother

\hypersetup{
colorlinks = true,
linkcolor = RoyalBlue,
anchorcolor = blue,
citecolor = Blue,
filecolor = cyan,
menucolor = ForestGreen,
runcolor = cyan,
urlcolor = RoyalBlue}

\include{math_commands.tex}

\usepackage{xspace}
\newcommand{\GA}[0]{\textcolor{BrickRed}{\texttt{GA}}\xspace}
\newcommand{\BASE}[0]{\textcolor{gray}{\texttt{Baseline}}\xspace}
\newcommand{\ICUL}[0]{\textcolor{Green}{\texttt{ICUL}}\xspace}
\newcommand{\RANDOM}[0]{\textcolor{Blue}{\texttt{Benchmark}}\xspace}

\usepackage[accepted]{icml2024}

\usepackage{amsmath}
\usepackage{amssymb}
\usepackage{mathtools}
\usepackage{amsthm}

\usepackage[textsize=tiny]{todonotes}

\icmltitlerunning{In-Context Unlearning: Language Models as Few-Shot Unlearners}

\begin{document}

\twocolumn[
\icmltitle{In-Context Unlearning: Language Models as Few-Shot Unlearners}



\icmlsetsymbol{equal}{*}

\begin{icmlauthorlist}
\icmlauthor{Martin Pawelczyk}{yyy}
\icmlauthor{Seth Neel}{equal,yyy}
\icmlauthor{Himabindu Lakkaraju}{equal,yyy}
\end{icmlauthorlist}

\icmlaffiliation{yyy}{Harvard University, US}

\icmlcorrespondingauthor{Martin Pawelczyk}{martin.pawelczyk.1@gmail.com}

\icmlkeywords{Machine Learning, In-Context Learning, Unlearning, In-Context Unlearning, Data Deletion, Copyright Infringement, Large Language Models (LLMs), ICML}

\vskip 0.3in
]



\printAffiliationsAndNotice{\icmlEqualContribution} 

\begin{abstract}

Machine unlearning, the study of efficiently removing the impact of specific training instances on a model, has garnered increased attention in recent years due to regulatory guidelines such as the \emph{Right to be Forgotten}. Achieving precise unlearning typically involves fully retraining the model and is computationally infeasible in case of very large models such as Large Language Models (LLMs). 
To this end, recent work has proposed several algorithms which approximate the removal of training data without retraining the model. These algorithms crucially rely on access to the model parameters in order to update them, an assumption that may not hold in practice due to computational constraints or having only query access to the LLMs. In this work, we propose a new class of unlearning methods for LLMs called ``In-Context Unlearning.'' This method unlearns instances from the model by simply providing specific kinds of inputs in context, without the need to update model parameters. To unlearn specific training instances, we present these instances to the LLMs at inference time along with labels that differ from their ground truth. Our experimental results demonstrate that in-context unlearning performs on par with, or in some cases outperforms other state-of-the-art methods that require access to model parameters, effectively removing the influence of specific instances on the model while preserving test accuracy.
\end{abstract}

\section{Introduction}  
\label{section:introduction}

Over the past decade, machine learning (ML) models have become ubiquitous in high-stakes decision making settings such as hiring, criminal justice, and credit scoring. To ensure responsible deployment and usage of these models in real-world applications, several regulatory guidelines have been introduced to protect user privacy~\citep{regulation2016regulation,ccpa2021}, one of which is called the \emph{Right to be Forgotten}. The Right to be Forgotten offers users more control over their personal data by allowing them to submit a deletion request to retract permission for a company to use their personal data at any time, even if, for example, the company has already trained an ML model with it~\citep{regulation2016regulation,ccpa2021,biega2020dm,goldsteen2021data}.
This raises a significant dilemma for organizations that aim to comply with the spirit of the regulation and avoid potentially breaking the law~\citep{voigt2017eu}, particularly concerning how exactly this data should be ``removed'' from any models trained on it. A second motivation comes from a concern orthogonal to privacy: copyright infringement. Generative models, like Large Language Models (LLMs), can often reproduce their training data verbatim or with only superficial changes. This can lead to credible claims of copyright infringement when the underlying data is protected by copyright. In such cases, when a copyrighted input to a model is generated, the model owner may be required to ``take down'' the copyrighted work from the model. Unlearning the corresponding input is one potential technical solution to comply with the take down request without retraining the model from scratch.  
\begin{figure}[tb]
\centering
\includegraphics[width=0.49\textwidth]{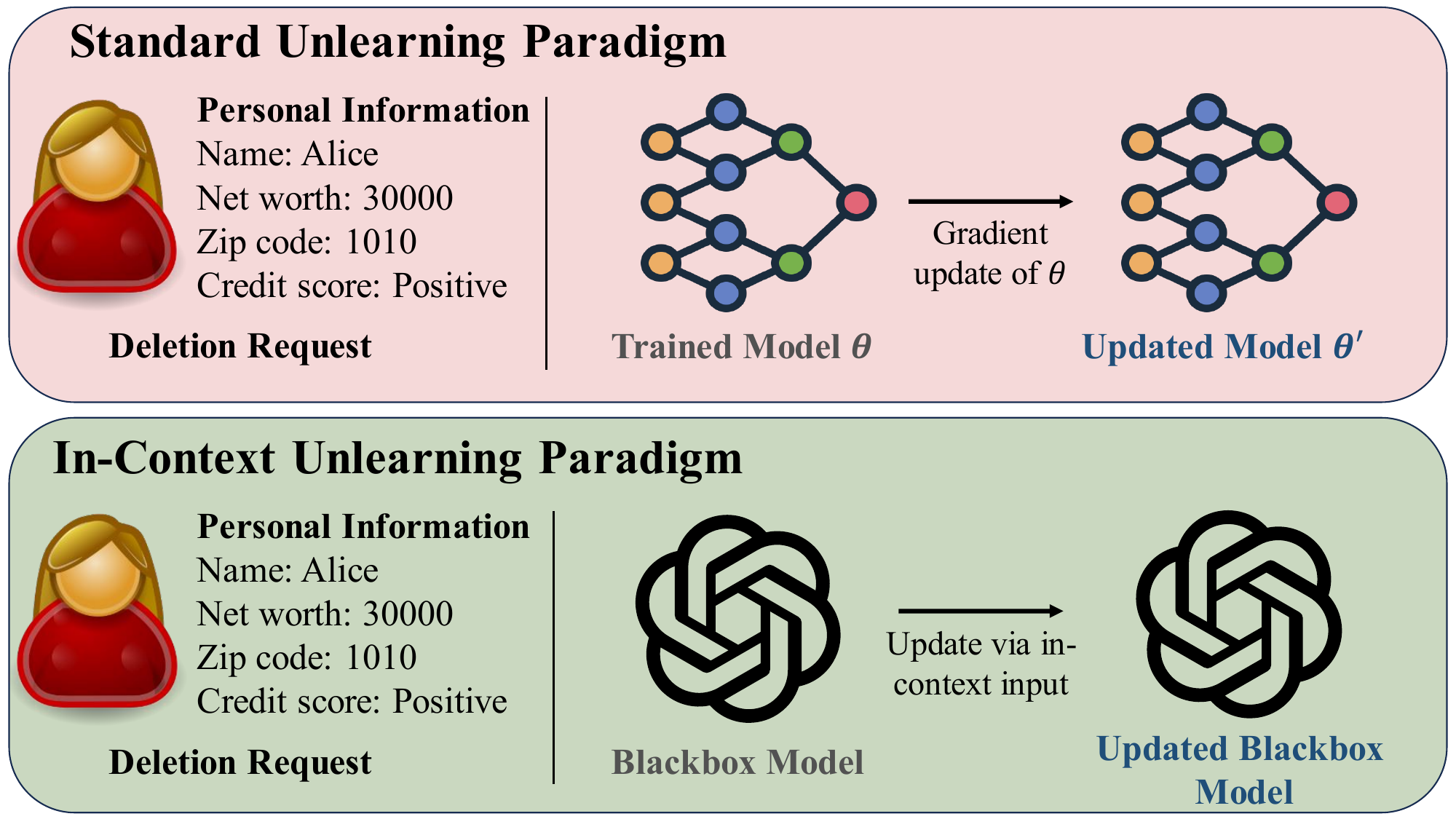}
\caption{
\textbf{Differences between In-Context Unlearning and Standard Unlearning}.
\textcolor{Tan}{{\textbf{Top}}}: Traditional unlearning approaches require access to model parameters $\theta$ and these parameters are updated in response to deletion requests. 
\textcolor{OliveGreen}{\textbf{Bottom}}: In-context unlearning does not require access to the parameters. Unlearning works by providing certain kinds of inputs in context which mimic the model’s performance as if the model was re-trained without the points.
}
\label{fig:teaser}
\vspace{-0.55cm}
\end{figure}
\\
\\
Indeed as a recent paper by \citet{henderson2023foundation} on copyright issues in generative models remarks:
\textit{retraining a model without a taken down datapoint could be exceedingly costly $\ldots$ new research is needed to identify new and improved mechanisms
for handling takedown requests in this relatively new setting.}
Work in ``Machine Unlearning'' seeks to bridge this gap by building algorithms that remove the influence of the deleted point from a trained model, while avoiding the computationally expensive step of fully re-training on the updated dataset \citep{ginart2019making, sekhari2021remember}. 
\begin{figure}[tb]
\centering
\resizebox{\columnwidth}{!}{
\begin{tcolorbox}[width=0.37\textwidth, nobeforeafter, colback={white}, title={Dataset used to finetune LLM}] 
\textbf{ID} \bm{$1$}: Name: Alice, Net Worth: 30K, Zip code: 1010. 
Score: Positive. \vspace{0.10cm}
 \\
\textbf{ID} \bm{$2$}: Name: Bob, Net Worth: 6K, Zip code: 1012.
Score: Neutral. \vspace{0.10cm} \\
$~~~~~~~~~~~~~~~~~~~~~~~~~~~ \vdots$
\vspace{0.10cm}
\\
\textbf{ID} \bm{$\text{N}$}: Name: Eve, Net Worth: -10K, Zip code: 0001.  
Score: Negative
\end{tcolorbox} 
\hfill 
\begin{tcolorbox}[width=0.38\textwidth, nobeforeafter, colback={white}, title={In-Context Input}] 
Name: Alice, Net Worth: 30K, Zip code: 1010. Score: \textcolor{Red}{Neutral}.
\vspace{0.10cm} \\
Name: Bob, Net Worth: 6K, Zip code: 1012. Score: Neutral.
\vspace{0.10cm} \\
Name: Eve, Net Worth: -10K, Zip code: 0001. Score: Negative
\vspace{0.10cm} \\
$~~~~~~~~~~~~~~~~~~~~~~~~~~~ \vdots$
\end{tcolorbox} 
}
\caption{\textbf{Demonstrating in-context unlearning}.
\textbf{Left}: The data set used to finetune the LLM. \textbf{Right}: In-context unlearning removes the influence that samples from the forget set $S_f$ (\textbf{ID} \bm{$1$} from the dataset) have on the completion by adding examples from the forget set with different labels to the in-context input (e.g., for ``\text{Name: Alice, Net Worth: 300K, Zip code: 1010}'' the label was changed randomly from Positive to \textcolor{Red}{Neutral}).}
\vspace{-0.60cm}
\label{fig:icul}
\end{figure}

At the same time as ML privacy regulation has started to gain traction, the release of Large Language Models (LLMs) has marked a pivotal transition in ML research \citep{brown2020language}. 
Modern LLMs have demonstrated competency in a vast array of challenging tasks, ranging from language comprehension \citep{radford2019language}, reasoning \citep{bubeck2023sparks} to tabular data generation \citep{borisov2022language}. 
These models not only exhibit effective abilities on tasks they were designed for, but they also display remarkable adaptability to unfamiliar tasks.
This surprising versatility is partially attributed to a learning paradigm called ``in-context learning'' \citep{brown2020language}, wherein the model has access to a set of in-context examples, a minimal collection of input and label pairs, that are added to the prompt at inference time to enhance LLM performance.



Despite the prominence of LLMs, and extensive recent work on machine unlearning, studying unlearning in LLMs is relatively unexplored \citep{jang2022knowledge}.
Perhaps this is because compared to conventional machine unlearning on image classifiers for example, unlearning in LLMs has two additional challenges.  
First, many LLMs operate as black-boxes 
(see Figure \ref{fig:teaser}), meaning that 
standard unlearning techniques that perform gradient ascent or descent on the model's parameters cannot be implemented \citep{neel2021deletion}. 
Second, even if the unlearning algorithm has ``white-box'' access (access to model parameters), performing gradient updates on LLMs with many billions of parameters every time an unlearning request comes in might be computationally infeasible.

To address these challenges, we propose a novel class of unlearning methods suitable for large language models (see Figure \ref{fig:teaser}).
To the best of our knowledge, this work is the first to suggest In-Context UnLearning (\texttt{ICUL}) which deploys a uniquely built context to eliminate the influence of a training point on the model output. 
In order to unlearn a particular training instance, the model context is constructed in such a way that the labels associated with training points targeted for deletion are flipped randomly, and both the training points and their flipped labels are provided at the beginning of the context alongside additional correctly labelled context examples sampled from the training data distribution (see Figure \ref{fig:icul}).
Our \texttt{ICUL} method does not require knowledge of the LLM's parameters, and yet manages to achieve performance levels that are competitive with or in some cases exceed the state-of-the-art LLM unlearning methods which require access to LLM parameters and involve expensive gradient computations \citep{jang2022knowledge}.

We experiment with multiple established real world datasets: AG-News, SST-2, SQUAD, and Amazon reviews to evaluate the effectiveness of our proposed unlearning method. Our results on text classification and question-answering tasks clearly demonstrate the efficacy of the proposed unlearning method, and highlight that it practically eliminates a training point's influence on the model output.
These results indicate the significant potential for unlearning training points from black-box models.
Our proposed methods and findings offer a new perspective on unlearning mechanisms in LLMs:
\begin{itemize}
\itemsep-0.05cm 
\item \textbf{New unlearning paradigm for LLMs}: This is the first work to use in-context learning for machine unlearning by specifically constructing contexts that induce model behavior that is indistinguishable from the behavior of a re-trained model.
\item \textbf{New empirical unlearning evaluation:} In Subsection~\ref{section:approximate_unlearning} we introduce \texttt{LiRA-Forget}, a new unlearning evaluation that adapts the LiRA MIA \citep{carlini2021membership} to the problem of evaluating unlearning. We note that a similar evaluation metric was introduced concurrently by \citet{kurmanji2023towards}, and subsequently studied by \citet{hayes2024inexact} under the name U-LiRA. We discuss these works more in Section \ref{section:related_work}. 
\item \textbf{Data deletion from blackbox models}: \texttt{ICUL} does not require access to model parameters and can be readily applied to blackbox models. This makes it a useful tool to patch a model until the model can be updated or a retrained version can be deployed at the next deployment phase.
Thus, it is complementary to existing white-box unlearning techniques which have higher computational burdens. 
\item \textbf{Lower memory requirements}:
Our method boasts lower memory requirements compared to state-of-the-art unlearning methods like Gradient Ascent (\texttt{GA}), especially as the size of the LLM increases. For instance, on Llama-2 7B, \texttt{ICUL} runs on a Tesla V100 GPU with 32GB of RAM, while running \texttt{GA} would require access to an A100 GPU with 80GB of RAM. This makes \texttt{ICUL} computationally feasible for LLMs with billions of parameters.
\end{itemize}


\section{Related Work}
\label{section:related_work}
This work is the first to leverage in-context learning for machine unlearning, and one of the first to study unlearning in language models. 
Below we discuss related works for each of these topics.

\textbf{In-Context Learning.}
Transformers form the foundation of contemporary LLM architectures. 
The reason behind their remarkable achievements is thought to involve a concept called ``in-context learning'' (ICL) \citep{brown2020language,dong2022survey,liu2023pre}.
This refers to their ability to adapt to new tasks flexibly by incorporating data provided in the context of the input sequence itself, rather than fine-tuning which explicitly updates weights.
Exploring the full capabilities of ICL remains an active area of research, with recent works trying to understand its potential better empirically by studying in-context example design \citep{garg2022can,liu2022makes,min2022rethinking,liu2023pre}.
In particular, some works consider the relevance of ground-truth labels for ICL and find mixed results; \citet{min2022rethinking} find that ground-truth labels have little impact on classification performance while the findings by \citet{wei2023larger} suggest that only larger scale LLMs can adopt their predictions to align with flipped label contexts.
While all these works study how learning can be facilitated through in-context examples, none of these works explore how unlearning can be achieved by designing in-context examples.

\textbf{Machine Unlearning.}
Motivated by GDPR's ``Right to be Forgotten" recent literature develops procedures for updating machine learning models to remove the impact of training on a subset of points \citep{ginart2019making,wu2020deltagrad,Golatkar_2020_CVPR,golatkar2020forget,izzo2021,neel2021deletion,sekhari2021remember,jang2022knowledge,huang2023tight,wang2023kga} or a subset of concepts \citep{ravfogel2022linear,ravfogel2022adversarial,belrose2023leace} without having to retrain the entire model from scratch. Unlearning algorithms fall into two camps: \emph{exact unlearning} approaches that redesign training in order to permit efficient re-training (e.g., \citet{ginart2019making,sekhari2021remember}) and approximate unlearning which merely approximates retraining (e.g., \citet{neel2021deletion,jang2022knowledge}). The latter approach has been likened to ``forgetting" \citep{graves2021amnesiac,tirumala2022memorization,jagielski2022measuring} which tracks whether machine learning models progressively unlearn samples during the course of training and
is typically quantitatively assessed by membership inference (MI) attack accuracy \citep{jagielski2022measuring}. For simple hypothesis classes such as linear regression \citep{cook1980characterizations,guo2019certified,izzo2021} or kernel methods \citep{zhang2021rethinking,pawelczyk2022trade}, tailored machine unlearning methods exist that make use of closed for solutions for the updated model. 

For the majority of non-convex models used in practice, in order to preserve accuracy the training routine is left untouched. 
As a result, standard approximate unlearning algorithms do not have theoretical guarantees, and so they must be evaluated empirically. 
Prior to this work, these empirical guarantees typically relied on heuristics like comparing the loss of unlearned points to the average validation loss \citep{jang2022knowledge}, or aggregate properties of the unlearned model like the test error, error on the forget set, or distribution of model confidences on unlearned and test points \citep{golatkar2020forget}.
Our work proposes a more principled empirical unlearning evaluation based on constructing an optimal membership inference attack (MIA) to distinguish unlearned points from test points, based on the LiRA MIA \cite{carlini2021membership}, which is the state of the art MIA. 
We note that \citet{kurmanji2023towards} concurrently propose a similar evaluation they call LiRA-for-unlearning, and subsequently \citet{hayes2024inexact} perform a detailed benchmarking of existing unlearning algorithms using this metric, as well as more heuristic unlearning metrics. 
Critically they find that evaluations that do not evaluate the unlearning of a specific point $\textbf{x}$ using an example-specific threshold, tend to over-estimate unlearning performance relative to LiRA-based techniques.

Prior research has mostly explored  unlearning from discriminative classifiers, generally vision models (e.g., \citet{Golatkar_2020_CVPR,goel2022towards}), where the aim often is to forget entire classes like ``cats'' or ``ships.'' 
These approaches typically update the model by starting at the model produced after training, and taking either gradient ascent steps on the deleted points \cite{jang2022knowledge} or gradient descent steps on the retained points \citep{neel2021deletion}, sometimes combining both approaches simultaneously and adding regularization \citep{kurmanji2023towards, jia2024model, foster2023fast}. 


\vspace{-0.10cm}
\section{Preliminaries}
\label{sec:prelims}
Here, we first discuss the generic formulations of in-context learning. 
We then discuss how to measure unlearning success empirically.

\subsection{In-Context Learning}
In-context learning has recently emerged as a new paradigm that allows auto-regressive language models to learn tasks using a few examples in the form of context demonstrations \citep{brown2020language}.
Here, we follow common practice \citep{brown2020language,dong2022survey,liu2023pre}, and consider the following definition of in-context learning: 
For a given pretrained language model $f_{\theta}$, a set of context demonstrations $D_{\text{context}}$ and a query input, the language model generates a sequence of tokens with a predefined length.
For example, when the model is used for text classification, it typically outputs one additional token as its prediction from a set of $C$ possible tokens where $C$ is usually large (e.g., for the Bloom model $C=250680$).
The context $D_{\text{context}}$ consists of an optional task instruction and $L$ demonstration examples; $D_{\text{context}}$ = $\{$$[\text{Instruction input}]$ $[\text{Example input 1}] ~ [\text{Label 1}]$, $\dots$ $[\text{Example input L}] ~ [\text{Label L}]$$\}$.
The prompt, which uses $D_{\text{context}}$ along with the query $[\text{Query Input}]$, is then provided as input for the language model prediction.
In-context learning has emerged as a way to improve a pretrained model's predictions without the need of costly finetuning of the model for a specific task.



\subsection{\texttt{LiRA-Forget}: Measuring  Unlearning}  
\label{section:approximate_unlearning}

We now define how we measure (approximate) unlearning.
Our unlearning notion is that of \citet{ginart2019making, neel2021deletion}, but adapts the metric of MI attack success to operationalize this definition \citep{goel2022towards,golatkar2021mixed}. 
Let $S \subset \mathcal{S}^*$ denote the training set, sampled from a distribution $\mathcal{D}$. Let $\mathcal{T}: \mathcal{S}^* \to \Theta$ be the (randomized) training algorithm that maps $S$ to a parameterized model $f_{\theta(S)}$.
Further define the forget set as the subset of points to be forgotten from the trained machine learning model denoted by $S_f \subset S$.
We define an unlearning procedure $\mathcal{U}$ that takes as input the model $f_{\theta(S)}$, the forget set $S_f$ of data samples that should be deleted, and the train set $S$ (and possibly some auxiliary information which we suppress), and outputs an updated model $\bar{f} \sim \mathcal{U}(f_{\theta(S)}, S, S_f)$.
Denote the probability law of the training algorithm on input $S$ by $\probdist$, the law of the exact re-training algorithm by $\probretrain$, and the law of the unlearning algorithm by $\probunlearn$.
As first formalized in \cite{ginart2019making}, the goal of an approximate unlearning algorithm is to achieve small $d(\probretrain, \probunlearn)$ for some distance measure between distributions $d$. 
Empirically verifying whether $d(\probretrain, \probunlearn)$ is small is difficult for two reasons: i) For computational reasons we do not have direct access to samples from $\probretrain$, and ii) even if we did these distributions are extremely high dimensional and cannot be compared efficiently. 

We address issue (i) by approximating the re-training distribution via sample-splitting (described in more detail in Appendix \ref{app:machine_unlearning}); by training multiple models on splits of the data that do not contain $S_f$, we can approximate samples from $\probretrain$. This approach is known as training ``shadow-models'' and has been employed for MI in \citet{shokri2017membership}. We address (ii) by re-formulating the problem of bounding $d(\probunlearn, \probretrain)$ as a hypothesis testing problem. Le Cam's Lemma (see Theorem $2.2$ in \citet{tsyb}) establishes a correspondence between $d(\probunlearn, \probretrain)$ and the ability of an optimal hypothesis test to distinguish $\probunlearn$ from $\probretrain$ based on a single sample. More specifically, we imagine a model $f$ is sampled from $\probunlearn$ with probability $1/2$ else from $\probretrain$ with probability $1/2$, and conduct a hypothesis test to determine which distribution $f$ came from:
\begin{equation}
\text{H}_0: f \sim \probretrain \text{ vs. } \text{H}_1: f \sim \probunlearn.
\label{eq:hypothesis_testing_problem}
\end{equation}
Rejecting the null hypothesis corresponds to inferring that $f$ was not from the re-training distribution. The Neyman-Pearson lemma \citep{neyman1933ix} asserts that the optimal hypothesis test at a predetermined false-positive rate involves thresholding the likelihood-ratio test statistic $\Lambda$.
As discussed, approximating the exact likelihood ratio statistic $\Lambda$ is intractable due to the high dimensionality of $f$, and so we follow recent work on MIAs, that instead takes the likelihood ratio with respect to the distribution of losses on the forget points $S_f$ for both models. 
This is closely related to the \texttt{LiRA} attack statistic proposed in \citet{carlini2021membership}, but differs critically in that the numerator considers the model produced by training on $S_f$ \emph{and then unlearning} via $\mathcal{U}$ rather than the model that results after training. 
We then define the $\liraforget$ statistic $\hat{\Lambda}$:
\begin{equation}
\hat{\Lambda} = \frac{\prod_{(\bx, \mathbf{y}) \in S_f}\probunlearn\big(\ell\big(f (\bx),\mathbf{y}\big)\big)}{\prod_{(\bx, \mathbf{y}) \in S_f}p_{S \backslash S_f}\big(\ell\big( f(\bx), \mathbf{y}\big)\big)},
\label{eq:approx_lrt}
\end{equation}
where $\ell$ denotes an appropriate loss function.
As in these recent works we approximate the univariate distributions on losses in the numerator and denominator of \eqref{eq:approx_lrt} via sample-splitting. 
Specifically we fine-tune models on sub-sampled datasetes that either contain or do not contain $S_f$. 
To approximate the numerator, on the datasets that do contain $S_f$, we run $\mathcal{U}$ to unlearn $S_f$, and then compute the updated model's loss on $S_f$. To approximate the denominator, we simply take the models that were not trained on $S_f$ and compute their losses on $S_f$. 

\textbf{Operationalizing \texttt{LiRA-Forget}.}
Operationalizing the likelihood ratio test from \eqref{eq:approx_lrt} requires access to the distribution of losses under the null and alternative hypotheses.
While analytical solutions are usually not available, we can readily get large samples from these two distributions.
In an ideal scenario, this entails that we would need to fit as many re-train models and unlearned models as possible for every forget set of interest.
Since this approach becomes computationally too burdensome, we use the following approximation. 
We adapt the sample splitting procedure first introduced by \citet{carlini2021membership} to forget sets with sizes $J=\{1, 5, 10, 20\}$, and approximate the distributions under $H_0$ and $H_1$ from equation \eqref{eq:approx_lrt}.
We train $K$ shadow models on random samples from the data distribution $\mathcal{D}$ so that a fraction $p$ of these models are trained on the forget set $S_f = \{(\bx_j, \mathbf{y}_j)\}_{j=1}^J$, and a fraction $(1-p)$ are not.
In particular, we train shadow models on $K=10$ subsets of $\mathcal{D}$ so that each forget set $S_f \in S$ appears in $K\cdot p$ subsets.
This approach has the advantage that the same $K$ shadow models can be used to estimate the likelihood-ratio test for all the forget sets.
Finally, we fit the parameters of two Gaussian distributions to the confidence scores of the retain models and the unlearned models on $S_f$. 

\section{Our Framework: In-Context Unlearning}
\label{section:method}
In this section, we describe our framework, In-Context Unlearning (\texttt{ICUL}), in detail. 
Recall that the main goal of our framework is to eliminate the need to re-train the model from scratch or to update the parameters of the model when unlearning specific training data points. 
Instead, at inference time, we construct a specific context which lets a language model trained on a classification or question-answering task a behave as if it had never seen the specific data point during training before.

To this end, our framework leverages both correctly labeled / answered as well as mislabeled / incorrectly answered examples to construct an in-context input which is provided as input to the LLM at inference time.
By changing the labels / answers on the points targeted for unlearning, our method diminishes the model's confidence specifically on these instances, aligning them more closely with the model's confidence in the case where the points were never part of the training set.
In particular, the label / answer changing operation in Step $(1)$ below aims to remove the influence a specific training point has on the model outcome. 
Since Step $(1)$ may cause the model to ``overcorrect'' on the forget points leading to decreased test accuracy and invalid unlearning, Step $(2)$ from below serves as an efficient way to dampen the effect of flipping the labels / answers of forget points.
More specifically, we suggest the following 3 step in-context input construction approach which we term \texttt{ICUL} and which we illustrate for a classification task: \vspace{-0.3cm}
\begin{itemize}
\itemsep-0.05cm 
\item[\textbf{1)}] 
\textbf{Change labels on forget points to different labels.} Given a deletion request of size $K$, we randomly flip the labels on the corresponding $K$ training points whose influence should be removed from the model resulting in the template: \emph{``$[\text{Forget Input 1}]$ $[\text{Different Label}] ~ \cdots ~ [\text{Forget Input K}] ~ [\text{Different Label}]$''}.
\item[\textbf{2)}]
\textbf{Add $L$ correctly labeled training points.}
We randomly sample $L$ labeled examples and add them to the template of step 1, resulting in the updated template: \emph{``$[\text{Forget Input 1}] ~ [\text{Different Label}] ~ \cdots ~ [\text{Forget Input K}]$ $[\text{Different Label}] ~ \backslash n ~ [\text{Input 1}] ~ [\text{Label 1}] \cdots [\text{Input L}]$ $[\text{Label L}]$''}.
\item[\textbf{3)}] 
\textbf{Prediction.} Finally, we add the query input to the template resulting in the final prompt \emph{``$[\text{Forget Input 1}] ~ [\text{Different Label}] ~ \cdots ~ [\text{Forget Input K}]$ $[\text{Different Label}] ~ \backslash n ~ [\text{Input 1}] [\text{Label 1}] \cdots [\text{Input L}]$ $[\text{Label L}]$ $[\text{Query Input}]$''} and let the model predict the next token using temperature $t=0$.
\end{itemize}
If the underlying task is question-answering, then we analogously design the contexts by flipping the answers for the forget-targeted samples to other random answers from the dataset.

\section{Empirical Evaluation}
\label{section:evaluation}
\definecolor{Gray}{gray}{0.92}
\newcolumntype{g}{>{\columncolor{Gray}}c}
\definecolor{maroon}{cmyk}{0,0.87,0.68,0.32}
We now present our empirical analysis. First, we empirically show that in-context unlearning is successful at unlearning information from a finetuned LLM in a forward pass.
In Section~\ref{section:efficacy_unlearning}, we show that the unlearned model maintains extremely competitive model performance when using \texttt{ICUL} especially with larger deletion requests of 10 or 20 points despite having no access to model parameters. 
In Section ~\ref{sec:sensitivity_model_size} we demonstrate that our method also works reliably for larger LLMs like Llama-2 (7B), and finally in Section ~\ref{section:sensitivity_analysis} we show a variety of ablation experiments that emphasize that our method works as intended.
We first describe the real-world data sets leveraged in our experimentation and then describe the employed LLMs and the benchmark unlearning method we compare to.

\textbf{Datasets.} We evaluate our in-context input constructions on 3 standard text classification tasks, 
Stanford Sentiment Treebank (SST2) \citep{socher2013recursive}, Amazon polarity, and AG-News \citep{zhang2015character}.
The SST-2 dataset is derived from Rotten Tomatoes reviews \citep{pang2005seeing} and the task is to predict whether a given sequence of text has a positive or negative sentiment.
We also use the Amazon polarity and the AG-News datasets which were originally introduced by \citet{zhang2015character}. 
For Amazon, the task is binary classification for whether a given review is positive (four or five stars) or negative (one or two stars). 
For the AG-News dataset the task consists of classifying news articles in one of four classes `sports', `business', `world' or `technology'.
We also provide experiments on the standard SQUAD dataset \citep{rajpurkar-etal-2016-squad}, which represents a question-answering task. 
In line with work on auditing privacy leakages \citep{carlini2022quantifying,shokri2017membership}, we randomly sub sampled smaller data sets of 25000 points from each of these datasets for finetuning.
We show the average results over 10 runs for all of our experimental settings and usually report $\pm 1$ standard deviation across these runs. 

\textbf{Large Language Models.} We conduct experiments on Bloom (560M, 1.1B, 3B) \citep{bloom2022} and Llama-2 (7B) \citep{touvron2023llama} LLMs.
We finetune these models on the classification datasets using the following template for each sample: \emph{``$[\text{Input}]$ $[\text{Label}]$''}. 
We use the standard causal cross-entropy loss with initial learning rate set to $5\cdot 10^{-5}$ which encourages the model to predict the next token correctly given a total vocabulary of $C$ possible tokens, where $C$ is usually large (e.g., for the Bloom model $C=250680$).
At test time, the models predict the next token from their vocabularies given a context and query.

\textbf{Prior Baselines.} We implement the only available baseline for unlearning in LLMs from \citet{jang2022knowledge};
they unlearn via gradient ascent on the forget set, which can be interpreted as maximizing instead of minimizing the loss on the forget points.
We follow their suggestion and set the learning rate to $5\cdot 10^{-5}$, use one epoch and do sequential unlearning where every point from the forget set is individually and sequentially unlearned using a constant learning rate schedule. 
Additionally, since a learning rate of $5\cdot 10^{-5}$ usually led to poor results, we did a search over different learning rates $\{ 5\cdot 10^{-5}, 3\cdot 10^{-5}, 1\cdot 10^{-5}, 5\cdot 10^{-6}\}$. 
In the main text, we report the most competitive results.
\begin{figure}[t]
\centering
\includegraphics[width=0.49\textwidth]{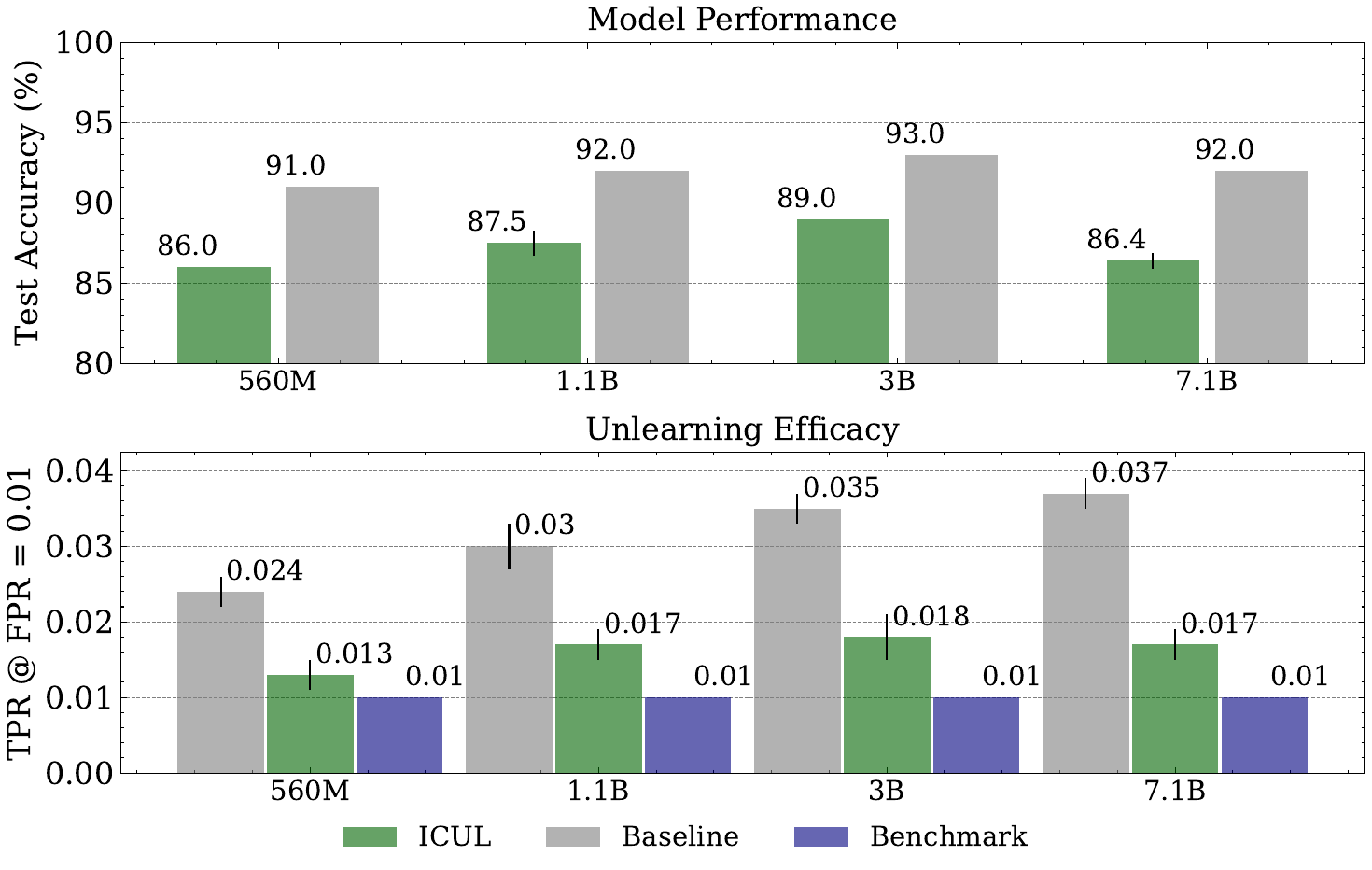}
\caption{\textbf{Evaluating \texttt{ICUL} sensitivity across model sizes.} We empirically assess unlearning via \textcolor{Green}{\texttt{ICUL}} with $L=6$ for 10 deletion requests on Bloom LLMs (560M, 1.1B, 3B, 7.1B) finetuned on the SST-2 dataset. 
\textcolor{gray}{\texttt{Baseline}} indicates performance when no unlearning is conducted, while \textcolor{Blue}{\texttt{Benchmark}} indicates best possible performance.
Vertical bars show $\pm 1$ standard deviation across 10 evaluation runs.}
\label{fig:llm_sizes}
\vspace{-0.5cm}
\end{figure}

\subsection{Evaluation Measures}
\label{sec:eval_measures}
When evaluating the efficacy of an unlearning method $\mathcal{U}$, three distinct objectives emerge:
validity of the unlearning procedure, classification accuracy post-unlearning, and run-time and space requirements of the algorithm. 
We first discuss measures that gauge unlearning validity.
We have described how to compute our unlearning success statistic $\hat{\Lambda}$, but it remains to discuss what values of $\hat{\Lambda}$ should be considered ``successful''.
We continue our analogy to recent work in evaluating membership inference attacks, and follow the paradigm introduced in \citep{carlini2021membership,leemann2023gaussian} that focuses on true positive rates (in this case of predicting that the loss came from the unlearned model) at low false positive rates as the most intuitive measure of MIA attack success. 
Unlike in the MIA context, where a successful attack has an AUC $\gg .5$, and an ROC curve that is above the diagonal even at very low FPRs, in our setting a successful unlearning algorithm corresponds to the failure of the LRT, and so we hope to see ROC curves that are very close to the diagonal even at low FPRs. 


\textbf{\texttt{Benchmark}: Random guessing performance.} 
The first measure consists of the decision not to unlearn the point from the model. 
It is represented by the dotted diagonal line indicating an equal ratio of FPR to TPR denoted as \texttt{Benchmark}.
For lower FPRs below $10^{-1}$, an unlearning method should demonstrate performance as close to the random guessing \texttt{Benchmark} as possible and below the \texttt{Baseline}, which we discuss next.

\textbf{\texttt{Baseline}: Train vs.\ held out samples on the initial model $\ftheta$}. 
This evaluation is a starting point measuring the initial information leakage from the model. 
It consists of the decision not to unlearn the point from the model and we will denote this as \texttt{Baseline} in all figures. 
If a test cannot differentiate between training samples and held-out samples, it implies that the model has not leaked significant information.
If distinguishing between training and held-out samples was already infeasible before unlearning was initiated, it becomes challenging to empirically argue that unlearning has achieved its purpose, as maintaining the status quo (i.e., doing nothing) would be a reasonable strategy. 
To conduct this evaluation, we run the \texttt{LiRA} attack using 10 shadow models \citep{carlini2021membership} on the model $\ftheta$. 

\begin{figure}[t]
\centering
\includegraphics[width=0.49\textwidth]{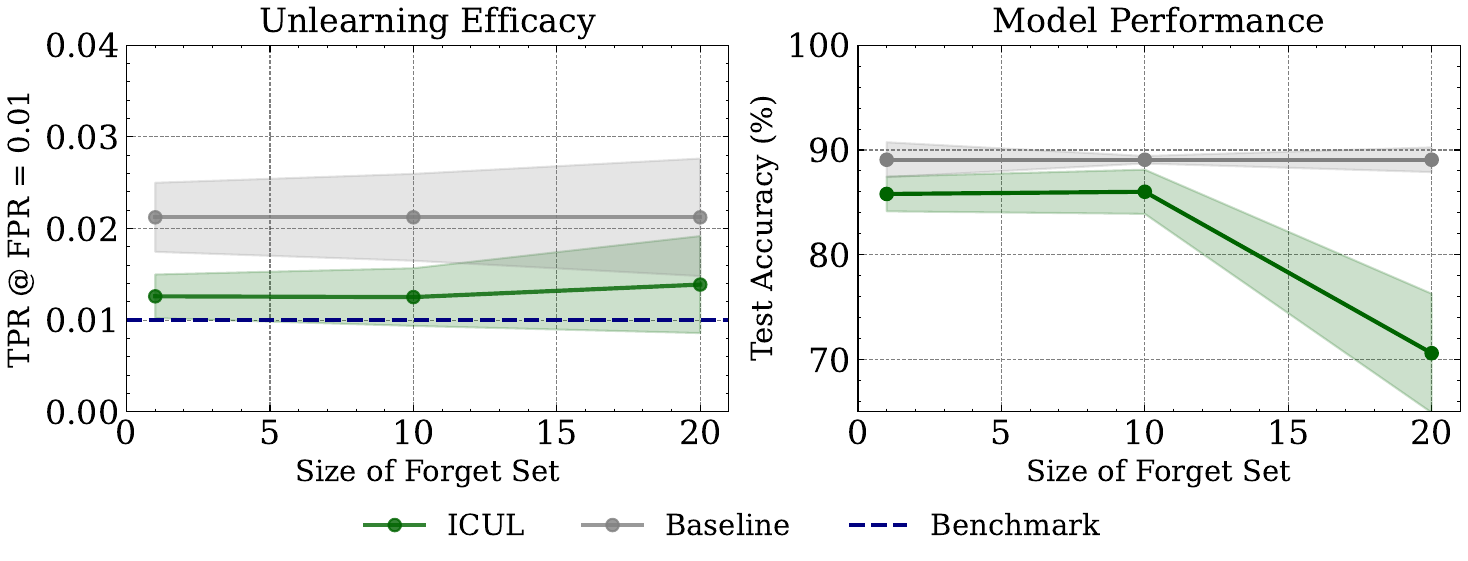}
\caption{\textbf{\texttt{ICUL} is effective on state-of-the-art LLMs.} 
We conduct empirical evaluations on unlearning via \textcolor{Green}{\texttt{ICUL}} with varying numbers of deletion requests $(1, 10, 20)$ for a Llama2 (7B) LLM fine-tuned on the SST-2 dataset. 
Finetuning required an A100 GPU (80 GB), while unlearning through \textcolor{Green}{\texttt{ICUL}} with $L=6$ was performed at inference time using a V100 GPU (32 GB).
\textcolor{gray}{\texttt{Baseline}} indicates performance when no unlearning is conducted, while \textcolor{Blue}{\texttt{Benchmark}} indicates best possible performance.
Shades indicate $\pm 1$ standard deviation across 10 evaluation runs.}
\label{fig:vary_ubs_llama}
\vspace{-0.4cm}
\end{figure}

\begin{figure*}[t]
\centering
\includegraphics[width=\textwidth]{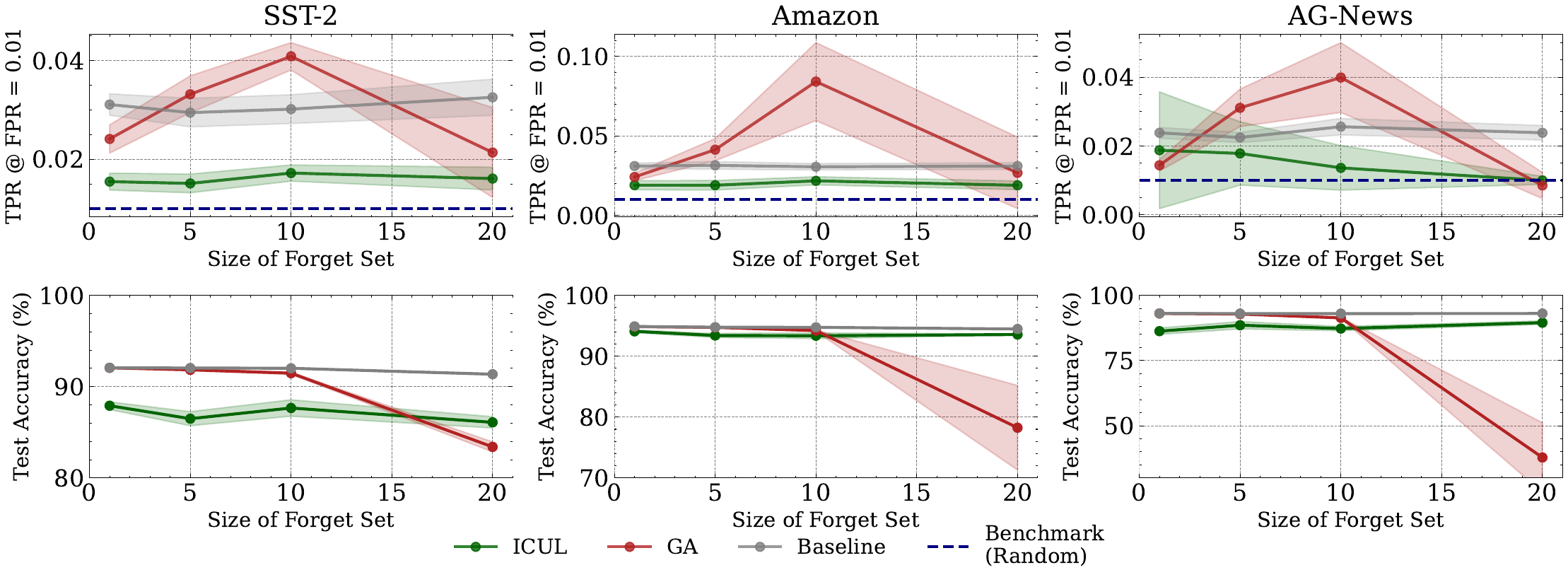}
\caption{\textbf{Evaluating unlearning.} We empirically evaluate unlearning across different unlearning methods for different number of deletion requests (1, 5, 10, 20) for a Bloom 1.1B model finetuned on different data sets (columns). 
For \textcolor{Green}{\texttt{ICUL}}, we select the most competitive results for $L \in \{2, 4, 6\}$, while for \textcolor{BrickRed}{\texttt{GA}} we search over learning rates of $\{5 \cdot 10^{-5}, 3 \cdot 10^{-5}, 1 \cdot 10^{-5}, 5 \cdot 10^{-6}\}$. Shades indicate $\pm 1$ standard deviation across 10 evaluation runs.
Reproducing these experiments requires approx.\ 1800 GPU hours on a V100 GPU (32GB).
\textbf{Top row -- Unlearning efficacy}: \texttt{LiRA-Forget} performance at a fixed FPR=$0.01$. 
\textcolor{gray}{\texttt{Baseline}} indicates performance when no unlearning is conducted.
Ideally, \GA and \ICUL performance curves trace significantly below the \BASE and as close to the random guessing \RANDOM (dashed line) as possible. \textbf{Bottom row -- Model Performance}: Test accuracies as we vary the number of deletion requests.}
\label{fig:vary_ubs}
\vspace{-0.4cm}
\end{figure*}

\textbf{Forget vs.\ held out samples on the updated model $\bar{f}$.} 
The key evaluation assesses the success of unlearning when the model is updated by either  \texttt{GA} or \texttt{ICUL}.
Can the model effectively forget the specific data point in question? 
I.e, is the model output on a data point when it is held out of the training set indistinguishable from the output on the same data point when it was initially part of the model but subsequently removed through the unlearning process? 
This critical evaluation is conducted by running our  \texttt{LiRA-Forget} attack against the model $\bar{f}$ as discussed in Section~\ref{sec:prelims}. 

\textbf{Evaluating model performance.} 
In addition to these evaluations, the overall performance of the model is a crucial consideration \citep{golatkar2021mixed}. 
The model's predictive capabilities should demonstrate effectiveness across various scenarios, including 1) train points $S$, 2) points $S_f$ targeted for unlearning and 3) randomly drawn test points.

\subsection{Empirically Evaluating Unlearning} 
\label{section:efficacy_unlearning}
In this Section, we evaluate the efficacy of unlearning various numbers of deletion requests $(1, 5, 10, 20)$ for a Bloom 1.1B model for all considered data sets. 
The results are summarized in Figure \ref{fig:vary_ubs}.
We compare \texttt{GA}, which has access to model parameters, with our proposed \texttt{ICUL} method, and compare their performance to the \texttt{Baseline} and the random guessing \texttt{Benchmark}.


Inspecting the top row of Figure \ref{fig:vary_ubs}, we find the \texttt{ICUL} curves, for all datasets and sizes of deletion requests, trace close to the \texttt{Benchmark} that represents a random guess probability of whether a point intended for removal is still part of the LLM. 
Also note that our method consistently surpasses the \textcolor{gray}{\texttt{Baseline}} in terms of TPRs at FPRs of 0.01 on all datasets and across all deletion requests. 
When we contrast \texttt{ICUL} with \texttt{GA}, \texttt{ICUL} consistently achieves superior (lower) TPRs at FPRs of 0.01, besting \texttt{GA} in 14 out of 16 cases.
Surprisingly, these results show conclusively that \texttt{ICUL} unlearns more effectively than \texttt{GA} using the strong \texttt{LiRA-Forget} evaluation, despite being moch more memory efficient (see App.\ \ref{app:add_results}), and requiring only black-box model access. 

Figure \ref{fig:vary_ubs} also reveals intriguing patterns regarding \texttt{GA}: for both small and large deletion requests, \texttt{GA} demonstrates reasonable performance, outperforming \texttt{Baseline}, but in the middle ground of 5 and 10 deletion requests performance drops below even the trivial baseline. 

\textbf{Evaluating the unlearned models' performance.}
Next, we assess the performance of the models post-unlearning, using accuracy as the evaluation metric.
An overview of these results can be found in the bottom row of Figure \ref{fig:vary_ubs}.
For results on the forget points' and train points' performance see Figure \ref{fig:all_accuracies} in the Appendix.
In the main text, we focus on performance on the test points. 
While \texttt{GA} exhibits better test accuracy than \texttt{ICUL}, as we expand the number of deletion requests to 10, the performance gap between \texttt{ICUL} and \texttt{GA} on unseen test data starts narrowing down.
Remarkably, for 20 deletion requests, the performance of \texttt{GA} drops significantly while \texttt{ICUL} maintains a similar level of test accuracy regardless of the number of deletion requests. 
Even below deletion requests of size 20, \texttt{ICUL} obtains reasonable test accuracy on all datasets, and on Amazon the test accuracy is within $1\%$ of the performance of the \textcolor{gray}{Baseline}.

\subsection{Sensitivity of In-Context Unlearning to Model Size}
\label{sec:sensitivity_model_size}

\textbf{Impact of varying model size on \texttt{ICUL}.}
When assessing \texttt{ICUL} across varying model sizes, two discernible trends emerge (refer to Figure \ref{fig:llm_sizes}). First, somewhat surprisingly, there is a positive correlation between model size and the percentage improvement over the \texttt{Baseline}: 41.67\% for the 560M LLM, 43.33\% for the 1.1B LLM, for the 3B LLM 51.42\%, and for the 7.1B LLM 54.05\% (see Figure \ref{fig:llm_sizes}, bottom). 
Second, we observe a relationship between the number of parameters in the LLM and the post-unlearning model performance, as evidenced by an increase in test accuracy from 86.1\% for the 560M LLM to 89\% for the 3B LLM (see Figure \ref{fig:llm_sizes}, top).

\begin{figure*}[htb!]
\centering
\includegraphics[width=\textwidth]{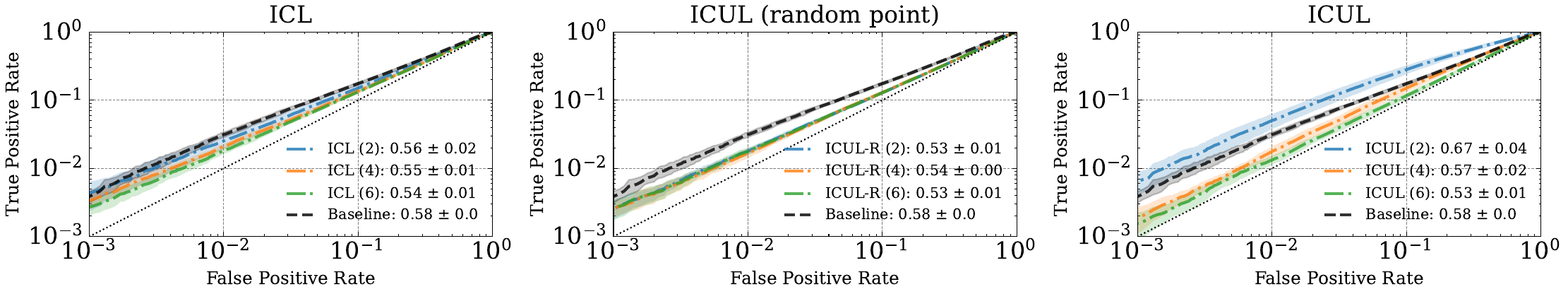}
\caption{\textbf{Towards understanding \texttt{ICUL}.} 
We plot \texttt{LiRA-Forget} performances using log-scaled AUC curves for a Bloom 1.1B LLM finedtuned on the SST-2 dataset. 
Here we consider 1 deletion request. 
The different experiments are described in more detail in Section \ref{section:sensitivity_analysis}. \textbf{Left}: Standard in-context learning with correct label. \textbf{Center}: Here we follow the \texttt{ICUL} construction, but where the forget point is exchanged for a random point from the data distribution. \textbf{Right}: Our suggested \texttt{ICUL} as described in Section \ref{section:method}. 
Closer proximity to the dotted diagonal, symbolizing the \texttt{Benchmark}, indicates superior performance.
}
\label{fig:ubs1_probing_context}
\vspace{-0.4cm}
\end{figure*}

\textbf{Exploring sensitivity across state-of-the-art LLMs.}
Here, we examine the sensitivity of \texttt{ICUL} results to the class of language models employed. 
To gauge this, we assess the performance of our unlearning algorithm in erasing 1, 10, and 20 points from the SST-2 dataset using the Llama-2 (7B) LLM, recognized as one of the top-tier models within the 7B parameter class of LLMs \citep{touvron2023llama}. 
The findings, depicted in Figure \ref{fig:vary_ubs_llama}, affirm that \texttt{ICUL} extends its effectiveness to other classes of LLMs, exhibiting forgetting efficacy comparable to that observed in Bloom models.
Notably, akin to the smaller Bloom models, the TPR at a FPR of 0.01 closely aligns with the \texttt{Benchmark} across all deletion request sizes.
\subsection{Broadening the Scope of In-Context Unlearning} \label{sec:broaden_scope}

We explore the possibility of using \texttt{ICUL} on additional language tasks, conducting empirical evaluations on a question answering task using the SQUAD dataset \citep{rajpurkar-etal-2016-squad}.
We focus on unlearning 5 and 10 samples from the finetuned model.
The results summarized in Table \ref{tab:performance_squad} demonstrate that \texttt{ICUL} is effective beyond multiclass classification tasks; \texttt{ICUL} reduces privacy leakage by 24\% for 5 deletion requests and by 43\% for 10 deletion requests compared to the \texttt{Baseline}, performing close to the \texttt{Benchmark}. 
While the accuracy drop is a moderate 5.5\% for 5 deletions, it becomes a significant 16.7\% for 10 deletions.

\begin{table}[tb]
\centering
\resizebox{\columnwidth}{!}{%
\begin{tabular}{lccc}
\toprule
 & Accuracy & TPR @ FPR = $10^{-2}$ & \texttt{Benchmark} \\
\cmidrule(lr){1-4}
\multicolumn{4}{c}{5 Deletions} \\
\cmidrule(lr){1-4}
\texttt{Baseline} & 72.0\% & 0.0241 & 0.01 \\
\texttt{ICUL}     & 68.7\% & 0.0183 & 0.01 \\
\cmidrule(lr){1-4}
 \multicolumn{4}{c}{10 Deletions} \\
\cmidrule(lr){1-4}
\texttt{Baseline} & 72.2\% & 0.0247 & 0.01 \\
\texttt{ICUL}     & 60.3\% & 0.0140 & 0.01 \\
\bottomrule
\end{tabular}
}
\caption{Comparison of performance metrics for \texttt{Baseline} and \texttt{ICUL} methods with 5 and 10 deletions when performing unlearning on the SQUAD dataset.}
\vspace{-0.80cm}
\label{tab:performance_squad}
\end{table}

\subsection{Towards Understanding In-Context Unlearning}
\label{section:sensitivity_analysis}
Next, we study the factors in the context construction that lead to successful in-context unlearning, conducting additional analyses where we vary the context length, label flipping, and role of the forget point in the \texttt{ICUL} context from Steps 1 and 2.

\textbf{Varying context length}.
One key factor to consider is the length of the context.
This might influence the unlearning process.
So, our framework considers a few different context lengths  by varying the total number of correctly labelled context examples $L \in \{2, 4, 6\}$, which we refer to as \texttt{ICUL(L)}.
For \texttt{ICUL}, changing the context length can significantly improve results as seen on the right in Figure \ref{fig:ubs1_probing_context}. With shorter context lengths, such as 2, the reversed label of the forget point typically leaves an overly negative impact on the model's confidence scores. 
This generally results in poorer average performance than the \texttt{Baseline}, as shown by the comparison of their AUC scores (e.g., \texttt{ICUL(2)} scores at 0.67 while \texttt{Baseline} at 0.58). Furthermore, context lengths of this size are often not sufficient enough to reduce TPRs at FPR levels of $\{10^{-3}, 10^{-2}, 10^{-1}\}$ down to the level of random guessing benchmark.
On the other hand, 4 or 6 additional context examples tend to yield the best performance.

\textbf{ICL.}
Another crucial consideration is examining the necessity of label flipping for successful unlearning, and including a baseline where we avoid label flipping of the point that should be unlearned from step 1, which results in the following in-context input:
\emph{``$[\text{Forget Input}]$ $[\text{Label}]$ $\backslash$n  $[\text{Input 1}]$ $[\text{Label 1}]$ $\backslash$n $\cdots$ $[\text{Input L}]$ $[\text{Label L}]$ $[\text{Query Input}]$''}.
We term this setting \texttt{ICL(L)} as it corresponds to standard in-context learning.
Here we empirically study the effect of label flipping on unlearning success.
A comparison of the standard \texttt{ICL} approach (Figure \ref{fig:ubs1_probing_context}, left), where the label of the point we aim to remove is kept unchanged, with our proposed \texttt{ICUL}method (Figure \ref{fig:ubs1_probing_context}, right) illustrates that label flipping is a crucial factor that pushes the \texttt{ICUL} curve closer to the random guessing benchmark. 
This finding highlights the essential role of label flipping in successful unlearning and complements recent studies that explore its significance in standard ICL \citep{min2022rethinking,wei2023larger}.
These studies suggest that only large-scale language models' test accuracy is affected by fully randomized label flipping.
Complementing these findings, our results suggest that smaller LLMs can adjust their predictions to mimic an output distribution that has never encountered the points aimed for removal before.




\textbf{Dependence on forget point.}
The last key aspect to consider is whether \texttt{ICUL} requires dependence on the points to be forgotten. 
To analyze this aspect, the unlearning point from step 1 is substituted with a randomly selected training point paired with a different label, resulting in the subsequent prompt: \emph{``$[\text{Random Train Input}]$ $[\text{Different Label}]$ $\backslash$n  $[\text{Input 1}]$ $[\text{Label 1}]$ $\backslash$n $\cdots$ $[\text{Input L}]$ $[\text{Label L}]$ $[\text{Query Input}]$''}. 
We call this setting  \texttt{Random ICUL(L)}.
Therefore, we examine whether the point intended for deletion needs to be part of the prompt. 
Evidence supporting this requirement is displayed by comparing the middle and right plots in Figure \ref{fig:ubs1_probing_context}.
This comparison highlights that in the low FPR regime at or below $10^{-2}$, our proposed \texttt{ICUL} method substantially surpasses the \texttt{ICUL} that uses a random point.

Taken together, these results show that it is not merely providing examples in-context that results in the measured unlearning, it is the fact that we specifically change the label of the point(s) in question, and then pad the context with 2 to 6 examples with the correct label.





\section{Conclusion}
\label{section:conclusion}
\looseness=-1 
In this work, we presented a novel class of unlearning algorithms for LLMs that unlearn even without access to the model parameters. Our method effectively creates a model output distribution that mimics the scenario where a particular point was never part of the model's training dataset. 
Our algorithm for \texttt{ICUL} creates prompts comprising data points targeted for removal, their changed labels, as well as other accurately labeled instances, which are then provided as inputs to the LLM during inference. In order to evaluate our unlearning algorithm, we extend prior work on membership inference and measuring forgetting to empirically measure unlearning using a likelihood-ratio based test we call \texttt{LiRA-Forget}.

Our empirical results suggest that \texttt{ICUL} reliably removes the influence of training points on the model since \texttt{LiRA-Forget} cannot reliably distinguish between held out points and training points that were subsequently unlearned from the model. 
Because of its practical appeal and the novelty of our approach, this work establishes a novel  perspective on the field of machine unlearning.

Finally, our work offers several questions for exploration:
\begin{itemize}
\setlength{\itemsep}{0.15cm}
\setlength{\parskip}{0pt}
\setlength{\parsep}{0pt}
\vspace{-0.35cm}
\item \textbf{Generalizing \texttt{ICUL} to more complex tasks:} The effective use and evaluation of \texttt{ICUL} for tasks such as open ended data generation remains unclear and requires further investigation.
\item \textbf{Enabling larger deletion requests:} The current context design makes handling larger deletion requests infeasible, presenting an opportunity for future work. 
\item \textbf{Reducing test time runtime:}
As deletion requests increase in size, our \texttt{ICUL} prompts become longer, which consequently increases the test time runtime. 
To address this, more sophisticated prompt strategies are needed to maintain unlearning efficacy without being dependent on prompt length.
\item \textbf{Improving prompt designs:}
Future research could explore savvier prompts to mitigate large accuracy drops.
In Section \ref{sec:broaden_scope}, we have demonstrated that \texttt{ICUL} can be adapted to question-answering tasks, but for larger deletion requests like 10, we observed that the model accuracy post unlearning dropped by more than 15\%. 
\item \textbf{Prompt inversion attacks against \texttt{ICUL}:} LLMs are vulnerable to various types of attacks, including extracting output layers from the logits, and reconstructing some of the prompts from the output logits, as demonstrated in recent research \citep{nasr2023scalable,morris2023language}.
In particular, \citet{morris2023language} suggest that in-context learning methods create new privacy risks; for example, it is possible to ``steal'' the prompt (and thus in-context examples) and leak personal information. 
An important question to consider is whether \texttt{ICUL} is susceptible to prompt stealing attacks, and if so, how \texttt{ICUL} can be modified to mitigate this risk.
\item \textbf{Developing more practical unlearning tests:}
In Section \ref{section:approximate_unlearning}, we introduced \texttt{LiRA-Forget}, a new likelihood ratio-based test for assessing unlearning success. 
This test requires training multiple shadow models, which can be computationally infeasible if the models are large and require significant GPU resources. Future research should focus on developing more efficient unlearning tests that can be more easily applied to (finetuned) large language models with billions of parameters. 
\end{itemize}

\section*{Acknowledgements}
This work is supported in part by the NSF awards IIS-2008461, IIS-2040989, IIS-2238714, the AI2050 program at Schmidt Sciences, and faculty research awards from Google, JPMorgan, Harvard Data Science Initiative, and the Digital, Data, and Design (D\^{}3) Institute at Harvard. The views expressed here are those of the authors and do not reflect the official policy or position of the funding agencies

\section*{Impact Statement}
Our proposed In-Context Unlearning method presents a valuable tool for decision makers seeking to make a finetuned model behave as if specific data points had been removed from an LLM.
Our method has applications across various domains relying on algorithmic decision-making using classification algorithms, including healthcare, education, insurance, credit scoring, recruitment, and criminal justice. 

Recognizing the potential of \texttt{ICUL} comes with a responsibility to understand its nuances. 
While offering unparalleled capabilities, like any unlearning algorithm, \texttt{ICUL} is not without limitations. 
Failures in unlearning may manifest under specific hyperparameter configurations, such as setting the number of correctly labeled examples ($L$) too low. 
To maximize \texttt{ICUL}'s efficacy, decision-makers must grasp both its strengths and weaknesses. 
Therefore, we advocate for the complementary use of our proposed \texttt{LiRA-Forget} test, providing a robust mechanism for validating unlearning outcomes.


In navigating the complex terrain of algorithmic decision-making, our work not only presents a breakthrough method but also equips practitioners with the insights and tools necessary for responsible and informed unlearning of specific information in LLM based classification and question answering tasks.


\bibliography{bibfile}
\bibliographystyle{icml2024}

\newpage
\appendix
\onecolumn


\section{Reproducibility Statement}\label{app:reproducibility}
\textbf{Overall compute requirements.}
To run all experiments\footnote{We release our code at: \url{https://github.com/MartinPawel/In-Context-Unlearning}.}, we need to save the model weights of the shadow models required for \texttt{LiRA-Forget}.
To reproduce all our experiments, set 1500 GB of storage aside. 
All experiments that use Bloom models are run using Nvidia Tesla V100 GPUs (32 GB RAM).
For model finetuning on the Llama2 7B LLM, we use one A100 GPU (80 GB RAM); note that during finetuning we update \emph{all} 7B model parameters. 
For unlearning via \texttt{ICUL}, we use Tesla V100 GPUs (32 GB RAM).

\textbf{Experimentwise compute requirements.} Next, we provide an overview on the number of GPU hours to reproduce our experiments.
\begin{enumerate}
\item \textbf{Evaluating unlearning} (Figure \ref{fig:vary_ubs}): For every dataset, we finetuned 10 LLMs for one epoch for every forget set size (1, 5, 10, 20). This is required to run \texttt{LiRA-Forget}.
Finetuning one model roughly takes 1 GPU hour; finetuning all models roughly takes 40 GPU hours per dataset. 
Including hyperparameter search, we ran the unlearning procedures on \emph{all} 25000 points. 
First, for \texttt{ICUL} we ran inference across 3 context length configurations across 40 models and each run took 2 hours on average. 
Across all three data sets, this amount to a total of 600 GPU hours. 
For \texttt{GA}, the situation was similar. 
We ran the unlearning procedure for 4 learning rate configurations across 40 models and each run took roughly 2 hours.
This amount to a total of 600 GPU hours.
In total, to reproduce this experiment requires 1800 GPU hours of compute, or roughly 75 GPU days.
\item \textbf{Evaluating \texttt{ICUL} sensitivity across model sizes} (Figure \ref{fig:llm_sizes}): For the SST-2 dataset and the forget set size of 10, we finetuned 10 LLMs per model size. This took roughly 5 hours for the smaller 560M models, 10 hours for the 1.1B model and 14 hours for the 3B model. For each of these model, we ran inference 10 times. 
For the smaller 560M models, inference took roughly 2 hours per model, while inference took approx.\ 3 hours for the 1.1B model and roughly 5 hours for the 3B model. 
In total, this experiment will roughly take 130 GPU hours to reproduce.
\item \textbf{\texttt{ICUL} demonstrates effectiveness on state-of-the-art LLMs} (Figure \ref{fig:vary_ubs_llama}): 
For the SST-2 dataset and the forget set size of 1, 10 and 20 we finetuned 10 LLMs per forget set size. 
Finetuning Llama-2 (7B) took roughly 48 hours per forget set size on one A100 GPU (80GB).
Performing 10 inference runs on the three models required another 8 hours per run on average on the V100 GPU (32 GB). In total, this experiment roughly required 385 GPU hours.   
\end{enumerate}

\section{Details on the Machine Unlearning Evaluation}
\label{app:machine_unlearning}


\textbf{Operationalizing the Likelihood-ratio Audit.}
Operationalizing the likelihood ratio test from \eqref{eq:approx_lrt} requires access to the distribution of losses under the null and alternative hypotheses.
While analytical solutions are usually not available, we can readily get large samples from these two distributions.
In an ideal scenario, this entails that we would need to fit as many re-train models and unlearned models as possible for every forget set of interest.
Since this approach becomes computationally too burdensome, we use the following approximation:

\textbf{Approximating the distributions under $H_0$ and $H_1$ from equation \eqref{eq:hypothesis_testing_problem}.} Here we adapt the sample splitting procedure first introduced by \citet{carlini2021membership} to forget sets with sizes $J=\{1, 5, 10, 20\}$.
We train $K$ shadow models on random samples from the data distribution $\mathcal{D}$ so that a fraction $p$ of these models are trained on the forget set $S_f = \{(\bx_j, \mathbf{y}_j)\}_{j=1}^J$, and a fraction $(1-p)$ are not.
In particular, we train shadow models on $K=10$ subsets of $\mathcal{D}$ so that each forget set $S_f \in D$ appears in $K\cdot p$ subsets.
This approach has the advantage that the same $K$ shadow models can be used to estimate the likelihood-ratio test for all the forget sets.
Finally, we fit the parameters of two Gaussian distributions to the confidence scores of the retain models and the unlearned models on $S_f$. 
Across all experiments, we use $p=0.5$.

\textbf{Model losses.}
Instead of using the actual losses, we follow \citet{carlini2021membership} and compute model confidences as $\phi(f(\bx),\mathbf{y}) = \log (f(\bx)_y) - \log( \sum_{y'} f(\bx)_{y'})$ which the authors show yields the strongest empirical attack performance.
This score compares the confidence the model assigns to the true class (e.g., `positive') with the confidences the model assigns to all other classes (i.e., all other words from the approximately 250680 dimensional vocabulary). 
The higher the score is the more confident the model is in the correct prediction.

\section{Additional Results}\label{app:add_results}

\subsection{Additional Empirical Comparisons: Compute Times and Memory Requirements}
To illustrate an extensive computational cost comparison between \texttt{GA} and \texttt{ICUL}, we present evaluate memory cost and storage requirements for a Llama2 (7B) LLM on the SST-2 dataset.
It is important to note that \texttt{ICUL} is specifically designed for GPU RAM-constrained compute environments.
While increasing GPU RAM to fine-tune may pose challenges for many users, they might be more willing to accept additional computation time to obtain desired results. 
To provide a more nuanced perspective, we identify three distinct dimensions of computational costs associated with any unlearning method:
\begin{itemize}
\item The memory requirement for executing the model update (see Table \ref{table:gpu_ram_utilization});
\item The computational time required to update the model (see Table \ref{table:wall_clock_time});
\item The computational time needed to perform inference runs using the updated model (see Table \ref{table:inference_run_times}).
\end{itemize}
Looking ahead, further advancements in transformer architectures such as \citep{ding2023longnet} are significantly reducing inference times by improving computational efficiency for larger context lengths. In particular, the work by \citet{ding2023longnet} boasts linear computation complexity in the sequence length without sacrificing performance. 
This development will likely further enhance the utility of \texttt{ICUL} as compute times for larger contexts decrease.

\begin{table}[tb]
\centering
\begin{subtable}{0.33\textwidth}
\centering
\begin{tabular}{ccc}
\toprule
\# Deletions & \texttt{GA} & \texttt{ICUL} \\
\cmidrule(lr){1-3}
1  & 64.80 & 29.90 \\
5  & 64.79 & 30.57 \\
10 & 68.91 & 31.92 \\
20 & 68.90 & 34.16 \\
\bottomrule
\end{tabular}
\caption{Maximum GPU RAM utilization measured in GB for \texttt{GA} and \texttt{ICUL} across different numbers of deletions.}
\label{table:gpu_ram_utilization}
\end{subtable}
~~~~~~~
\begin{subtable}{0.33\textwidth}
\centering
\begin{tabular}{ccc}
\toprule
\# Deletions & \texttt{GA} & \texttt{ICUL}\\
\cmidrule(lr){1-3}
1  & 1.44 & 0.00 \\
5  & 3.41 & 0.00 \\
10 & 7.60 & 0.00 \\
20 & 15.94 & 0.00 \\
\bottomrule
\end{tabular}
\caption{Model update wall clock time measured in seconds for \texttt{GA} and \texttt{ICUL} across different numbers of deletions.}
\label{table:wall_clock_time}
\end{subtable}
\caption{Computational resources required to update Llama2 (7B) on the SST-2 dataset across unlearning methods.}
\end{table}

\begin{table}[tb]
\centering
\begin{tabular}{cccccc}
\toprule
\# Inference Runs & \texttt{GA} & \texttt{ICUL} (1 Deletion) & \texttt{ICUL} (5 Deletions) & \texttt{ICUL} (10 Deletions) & \texttt{ICUL} (20 Deletions) \\
\cmidrule{1-6}
1  & 0.40 & 0.42 & 0.57 & 0.99 & 1.64 \\
5  & 2.01 & 2.10 & 2.85 & 5.00 & 8.20 \\
10 & 4.03 & 4.20 & 5.70 & 10.00 & 16.40 \\
20 & 8.00 & 8.40 & 11.40 & 20.00 & 32.80 \\
\bottomrule
\end{tabular}
\caption{Inference run times measured in seconds for \texttt{GA} and \texttt{ICUL} across different numbers of deletions.}
\label{table:inference_run_times}
\end{table}

\subsection{Additional Empirical Comparisons: Test Accuracy and  Unlearning Efficacy}

Here we additionally experiment with the Yelp polarity data set that was was originally introduced by \citet{zhang2015character}.

\begin{figure*}[htb!]
\centering
\begin{subfigure}{\textwidth}
\includegraphics[width=\textwidth]{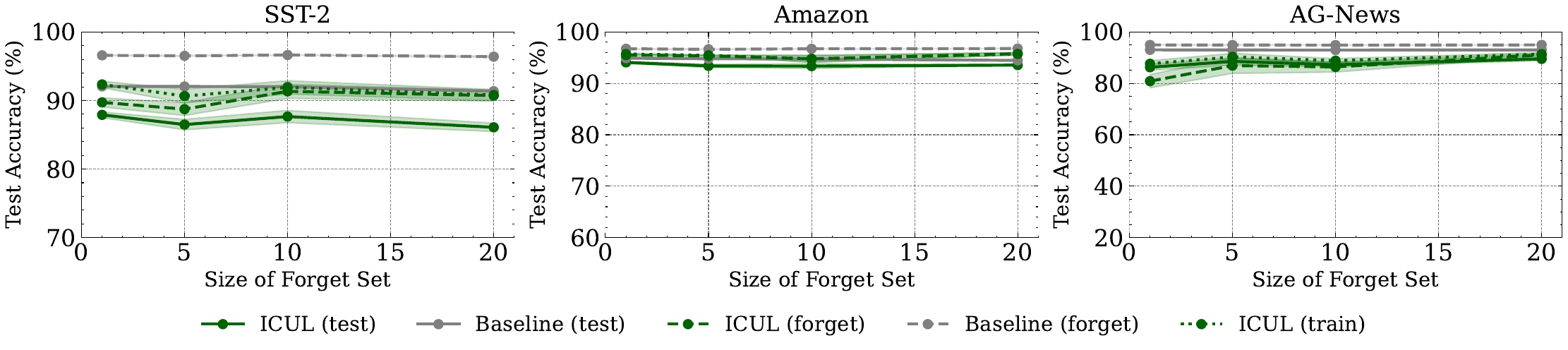}
\caption{\texttt{ICUL}}
\label{fig:ubs1_vary_context_bloom560}
\end{subfigure}
\vfill
\begin{subfigure}{\textwidth}
\centering
\includegraphics[width=\textwidth]{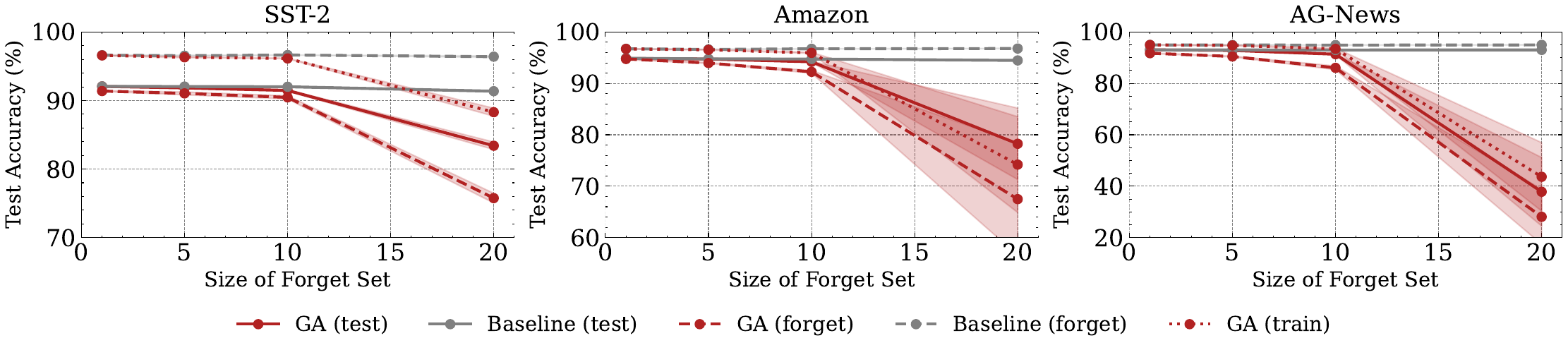}
\caption{\texttt{GA}}
\end{subfigure}
\caption{\textbf{Classification performance as we vary the size of the forget set}. We report classification accuracy on train, forget and test points across all data sets for the 1.1B Bloom LLM.}
\label{fig:all_accuracies}
\end{figure*}

\begin{figure*}[tb]
\centering
\begin{subfigure}{\textwidth}
\includegraphics[width=\textwidth]{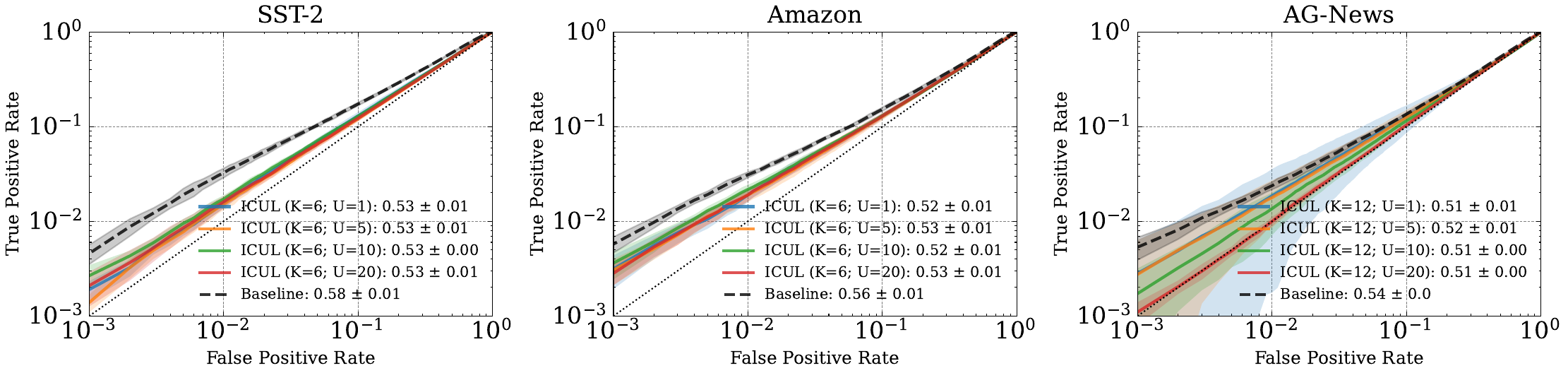}
\caption{\texttt{ICUL}}
\label{fig:}
\end{subfigure}
\vfill
\begin{subfigure}{\textwidth}
\centering
\includegraphics[width=\textwidth]{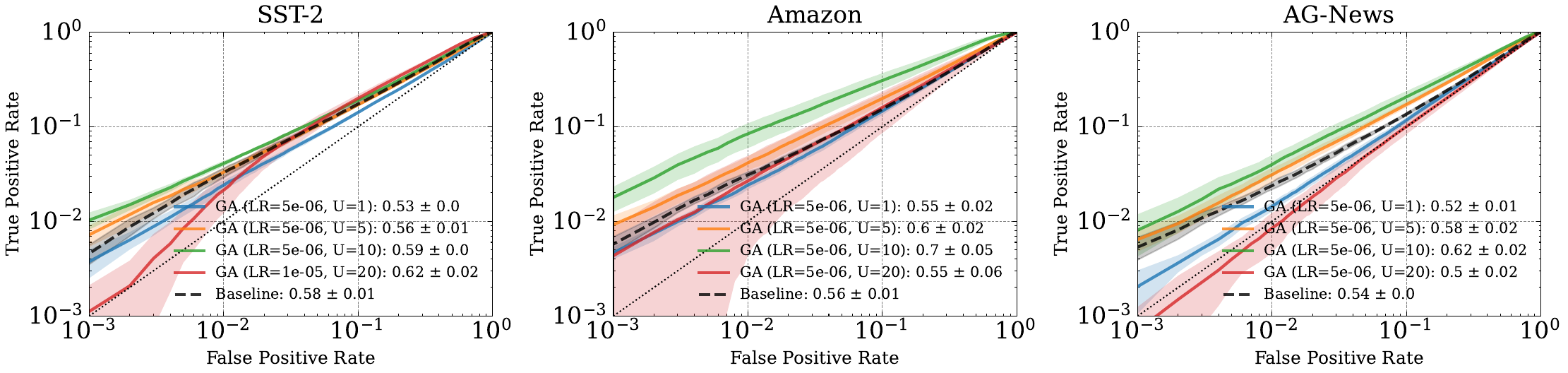}
\caption{\texttt{GA}}
\end{subfigure}
\caption{\textbf{Log scaled AUC curves}. Here we show the complete AUC curves for the most competitive hyperparameters on all data sets for the Bloom 1.1B model for both \texttt{GA} and \texttt{ICUL}.}
\label{fig:log_log_tprs}
\end{figure*}

\begin{figure}[tb]
\centering
\begin{subfigure}{0.90\textwidth}
\centering
\includegraphics[width=\textwidth]{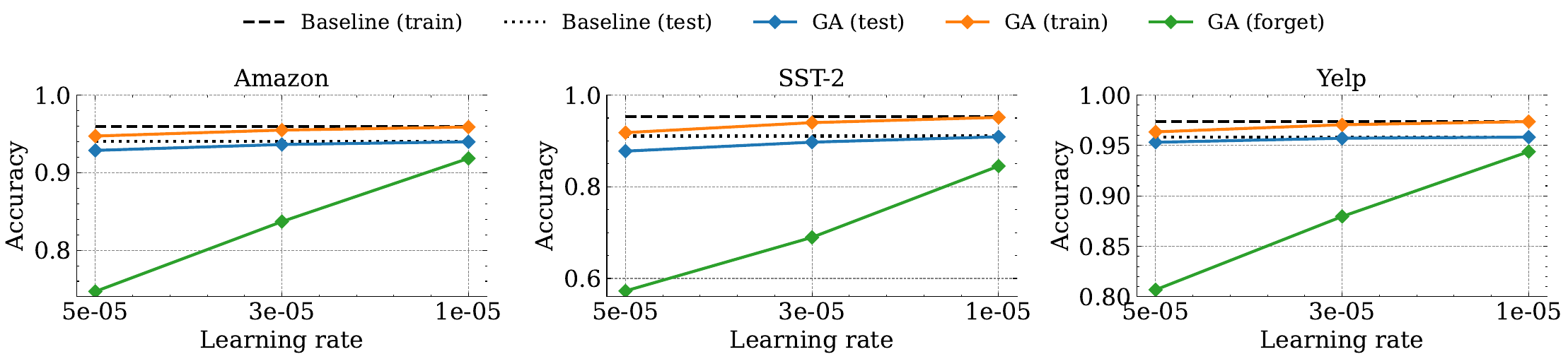}
\caption{Bloom (560M parameters)}
\label{fig:ubs1_performance_vary_lr_bloom560}
\end{subfigure}
\vfill
\begin{subfigure}{0.90\textwidth}
\centering
\includegraphics[width=\textwidth]{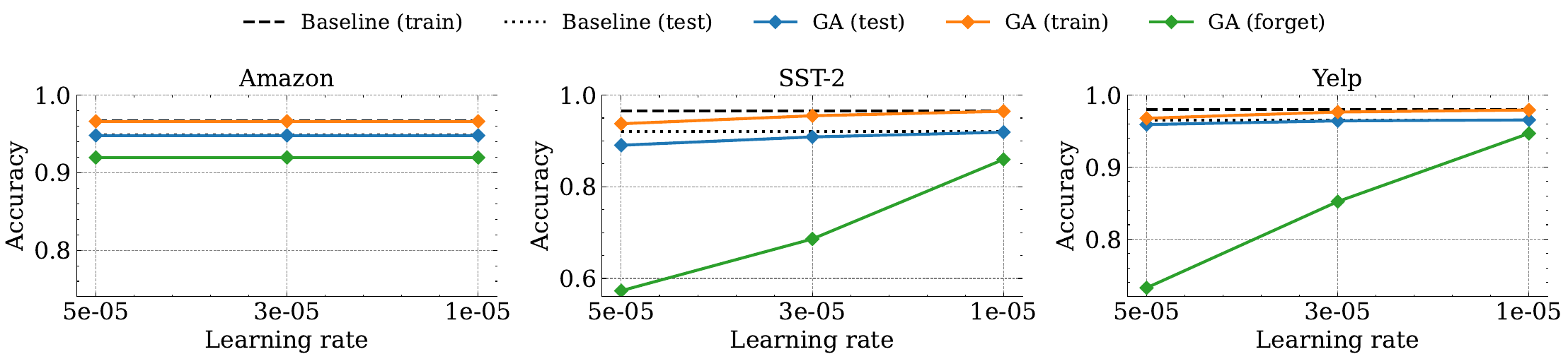}
\caption{Bloom (1.1B parameters)}
\end{subfigure}
\caption{\textbf{Classification performance as we vary the learning rate for \texttt{GA}}. We report classification accuracy on train, forget and test points across all data sets and model sizes.
For better readability, $\pm 1$ standard deviation was excluded from the figure.
}
\label{fig:ubs1_performance_vary_lr}
\vspace{-0.5cm}
\end{figure}
\begin{figure*}[tb]
\centering
\begin{subfigure}{0.90\textwidth}
\centering
\includegraphics[width=\textwidth]{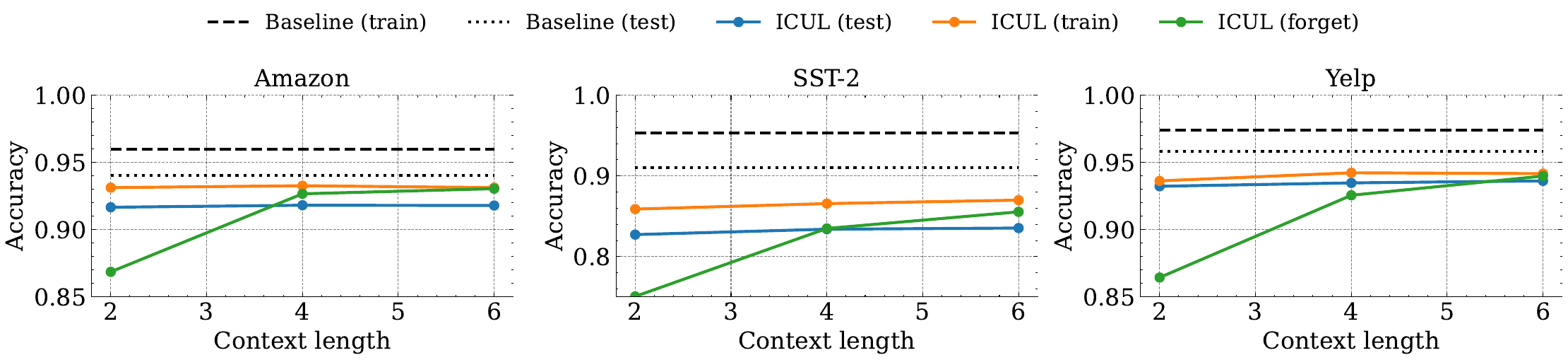}
\caption{Bloom (560M parameters)}
\label{fig:ubs1_performance_vary_context_bloom560}
\end{subfigure}
\vfill
\begin{subfigure}{0.90\textwidth}
\centering
\includegraphics[width=\textwidth]{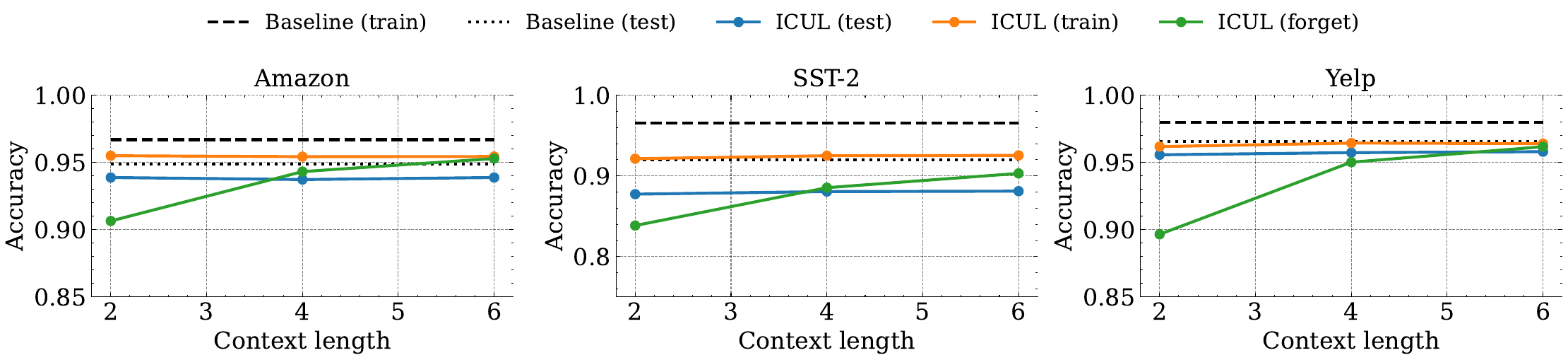}
\caption{Bloom (1.1B parameters)}
\end{subfigure}
\caption{\textbf{Classification performance as we vary context length for \texttt{ICUL}}. We report classification accuracy on train, forget and test points across all data sets and model sizes.
For better readability, $\pm 1$ standard deviation was excluded.
}
\label{fig:ubs1_performance_vary_context_length}
\end{figure*}

\end{document}